# NF-SAVO: Neuro-Fuzzy system for Arabic Video OCR

Mohamed Ben Halima, Hichem karray,
Adel. M. Alimi
REGIM: REsearch Group on Intelligent Machines
University of Sfax, National School of Engineers (ENIS)
B.P 1173. 3038 Sfax-Tunisia

Ana Fernández Vila
Grupo SSI, Dept. Ingeniería Telemática
University of Vigo
Spain

*Abstract*— **In this paper we propose a robust approach for text extraction and recognition from video clips which is called Neuro-Fuzzy system for Arabic Video OCR. In Arabic video text recognition, a number of noise components provide the text relatively more complicated to separate from the background. Further, the characters can be moving or presented in a diversity of colors, sizes and fonts that are not uniform. Added to this, is the fact that the background is usually moving making text extraction a more intricate process.**
**Video include two kinds of text, scene text and artificial text. Scene text is usually text that becomes part of the scene itself as it is recorded at the time of filming the scene. But artificial text is produced separately and away from the scene and is laid over it at a later stage or during the post processing time. The emergence of artificial text is consequently vigilantly directed. This type of text carries with it important information that helps in video referencing, indexing and retrieval.**

## I. INTRODUCTION

Optical Character Recognition for Arabic scripts is almost a solved problem for document images and researchers are now focusing on extraction and recognition of Arabic text from video scenes. This new and promising field in character recognition is called Arabic Video OCR and has several applications similar to video annotation, indexing, retrieval, search, digital libraries, and lecture video indexing. Images extracted from video sequences are of low resolution. For this, the commercial OCR provides poor results. For example, both Sakhr and OmniPage performed poorly on this image. Sakhr achieved 38.88% accuracy and OmniPage achieved 35.79% accuracy.

Works on text extraction may be generally grouped into four categories [11]: First category is the connected component methods which detect text by extracting the connected components of monotone colors that obey certain size, shape, and spatial alignment constraints. The second is the texture methods which treat the text region as a special type of texture and employ conventional texture classification method to extract text. The third is the edge detection methods which have been increasingly used for caption extraction due to the rich edge concentration in characters [8]. Finally, the correlation based methods which use any kind of correlation in order to decide if a pixel belongs to a character or not. Recognition of text in video is more difficult than many other OCR applications (e.g., reading printed matter) because of degradation, such as background noise, and deformation, like the variation in fonts.

Our system is divided into two steps: first we extract the textual information from the video sequence and second we recognize this text. Text recognition, even when applied to lines potentially containing text, remains a difficult problem given the variety of fonts and colors and the presence of complex background. Recognition is addressed by a segmentation step followed by a step of optical character recognition in a framework of multiple hypotheses.

In Figure 1, we present some examples of text in video frames.

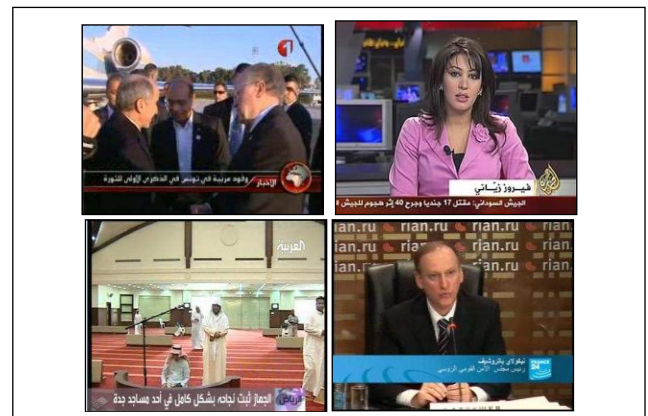

Figure 1. Examples of text in video frames

News programs are particular audio-visual documents they are generally formed by a set of semantically independent stories. For this reason before starting the extraction of the textual information from the news programs, segmentation into stories will be done. The following figure shows an example of image taken from video broadcast of Aljazeera TV, Tunisia 1 TV, France 24 TV and Alarabiya TV.

The rest of the paper will be organized in four sections. In the section 2, we discuss works related to news segmentation. In section 3, we will present how we detect and localize the textual information in video sequence. In section 4, we will present how we extract the text. In section 5, we will present how we recognize it. We conclude with directions for future work.

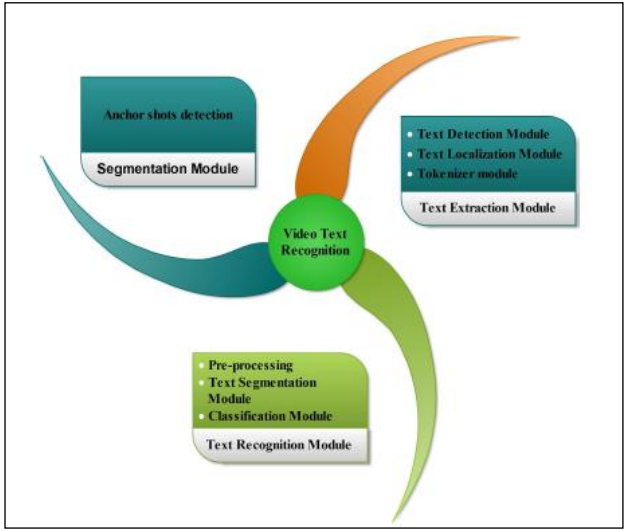

Figure 2.  Global Overview of our System

## II. SEGMENTATION MODULE

Stories segmentation is an essential step in any work done on video sequences. The story in every news program is the most important semantic unit. Story segmentation is an active research field with a lot of categories of works [16]. For this reason we find a special session concerning video stories segmentation in TRECVID campaigns [15] (TREC Video Retrieval Evaluation) which are the most challenging evaluation for video retrieval in the world. However, the majority of proposed works are based on detecting anchor shots. Indeed, the anchor shot is the only repetitive shot in a news program. The first anchor shot detector dates back to 1995 and it was proposed by Zhang [22]. In these works the anchor shot detection is based on classifying shots according to anchorperson shot models. The drawback of this method is the fact that it is impossible to define a standard model for all channels.

Now, we speak about multimodal anchor shot detection pioneered by Informedia. In this project Yang et al. [21] propose to use high level information (speech, text transcript, and facial information) to classify persons appearing in the news program into three types: anchor, reporter, or person involving in a news event. This method has been proved effective on TRECVID dataset. However, analyzing different video modalities including speech, transcript text, video frames and combining them to extract stories can take a lot of time.

In our system, video sequences are segmented by detecting and classifying faces to find group of anchor shots. This module is based on the assumption that the anchors faces are the only repetitive face throughout the entire program. The features that we extracted from faces will be used to cluster shots. The clustering technique that we used is the Kohonen map. They are well suited for mapping high dimensional vectors (shots) into two dimensional spaces. In fact, the Kohonen map is composed of two layers. The input layer corresponds to the input elements. The output layer corresponds to a set of connected neurons. Every input element is represented by a n-dimensional vector X = (x1, x2, …, xn) and connected to m nodes of the output layer through weights Wij (Figure 3).

After training the Kohonen map, we cluster the nodes of the Kohonen map in order to extract the region (cluster) in which the anchor shot are located. For this reason we used the CMeans classifier. However, the problem is that the C-Means classifier is the fact that it depedns on the number of clusters. It should be provided with the number of clusters to achieve the clustering.

Davies and Bouldin [7], solved this problem by proposing a progressive clustering ( 2 classes, 3 classes,…, N-max classes). For every clustering they compute a validity index which measures the quality of the clustering. The better classification has the lower of validity index, (Figure 4).

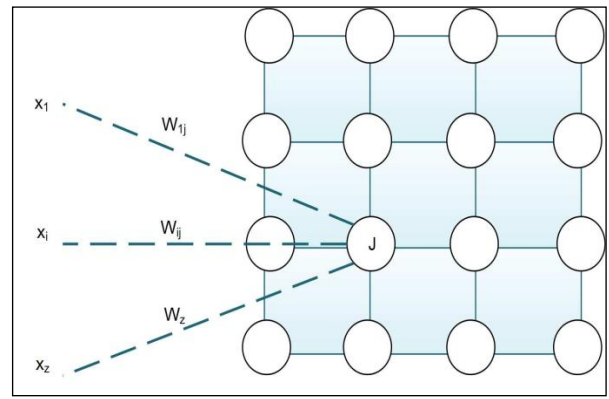

Figure 3.  Kohonen Map

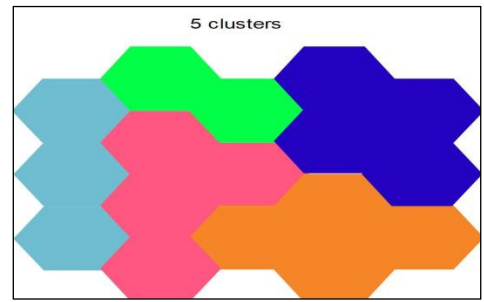

Figure 4.  Kohonen map after clustering

After clustering the Kohonen map we have to extract the cluster of anchor shots. For this reason we should consider these two assumptions. First the cluster of anchor shots should not contain less than for shots (faces), since a news story is made up by a minimum of four stories. Second, as the cluster of anchor shots is made by presenter faces only, so the density of this cluster is the smallest one, i.e. the average distance between the faces of the cluster. It is computed as follows:

$$MoyDis(C) = (\sum_{i=1}^{n}\sum_{j=i+1}^{n} \|X_i - X_j\|)/(n(n-1)/2) \quad (1)$$

## III. TEXT EXTRACTION

Generally, text embedded in video sequences can be classified into two categories: Text that is part of the scene and artificial text. Scene text (such as a sign outside a place of business or placards in front of conference participants) is part of image and usually does not represent information about image content, but artificial text (such as superimposed captions in broadcast news programs and other commercially produced videos) is laid over the image in a later stage. Artificial text is often good key for successful indexing and retrieval of videos. In news broadcast, overlaid text presents the circumstances as places or countries (Iraq, Israel, United States, etc.), gives the names of the interviewed persons or presents an important event (Olympic Games, hostage crisis, etc.). To extract text from stories shots we based on our work proposed in [16].

In this work we use an hybrid approach [14], which combines color and edges to extract text, proceeds as follows: text detection, localization and segmentation. Firstly, for text detection, we apply a new multiple frames integration (MFI) method to minimize the variation of the background of the video frames.

Thereafter, we eliminate from frame columns and rows of pixels which do not contain text. Secondly, we localize text pixels from the remaining rows and column clusters. Every window is represented by two frames. One is the frame of the window filtered along rows and the other one is the frame which is filtered along the columns. For every frame we achieve two operations: First, we realize a transformation from the RGB space to HSV space. Second, we generate, using Sobel filters, an edge picture. For every cluster of these frames, however, we formulate a vector composed of ten features: five of them represent the HSV image and the others represent the edge picture. These features are computed as follows: mean second order moment, third order moment, minimum value of the confidence interval and maximum value of the confidence interval.

For every cluster of these frames, we formulate a vector composed of ten features: five representing the HSV image and five representing the edge picture. These features are computed as follows: mean second order moment, third order moment, minimum value of the confidence interval and maximum value of the confidence interval.

The generated vectors will be presented to a trained back propagation neural network containing ten input nodes, 3 hidden ones and an output node .The training database contains 2000 key frames with the dimension of 320x240. The results of the classifications are two images: an image containing rows considered as text rows and an image containing columns considered as text columns. Finally, we merge results of the two images to generate an image containing zones of text. Once localized, the text in frame will be segmented. The segmentation process is computed as follows:

- Compute the Gray levels image.
- For each pixel in the text area, create a vector composed of two features: the standard deviation and the entropy of the 8 neighborhoods of pixels.
- Run the fuzzy C means clustering algorithm to classify the pixels into "text" cluster and "background" cluster.
- Binarize the text image by marking text pixels in black as shown in (Figure 5).

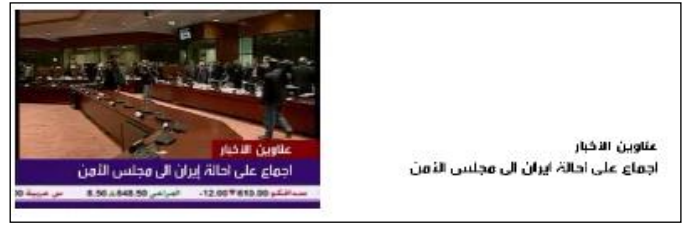

a) Original frame          b) Segmented text frame

Figure 5. Example of text extraction

## IV. ARABIC TEXT RECOGNITION

Many successful systems are found for the Latin or Chinese Video Text recognition, but few Arabic Video Text recognition systems are found. This is due to the standard features of Latin or Chinese text that facilitate the recognition process. For example, the Latin character shape do not changes when the character position changes in within a word; characters have only two shapes: capital and small. In opposing to Latin text, Arabic text is cursive and Arabic characters can have four diverse shapes because of their position within the word (Table 1).

Text recognition, even from the detected text lines, remains a challenging problem due to the variety of fonts, colors, presence of complex backgrounds and the short length of the text strings. An optical character recognition methodology was implemented including five successive stages: Pre-processing, Segmentation, Feature extraction, Classification and Post-processing.

### A. Morphological Structure of the Arabic Script

The alphabet from the Arabic language has 28 consonants (Table 1) including 15 from one to three points that differentiate between similar characters [Biadsy, 11]. The points and Hamza (ء) called secondary characters (complementary). They are located above the primary character as the "alif" (أ) below as the "Ba" (ب), or in the middle as the "jeem" (ج). There are four characters, which can take the secondary nature Hamzah (ء): alif (أ), waw (ؤ) kaf (ك) and ya (ئ) .

A distinguishing feature of Arabic writing is the presence of a base-line. The baseline is a horizontal line that runs through the connected portions of text (i.e. where the character's connection segments are located). The baseline has the maximum number of text pixels. Figure 6 demonstrates some of these characteristics on an Arabic sentence *[Mozaffari, 2008] [3]*.

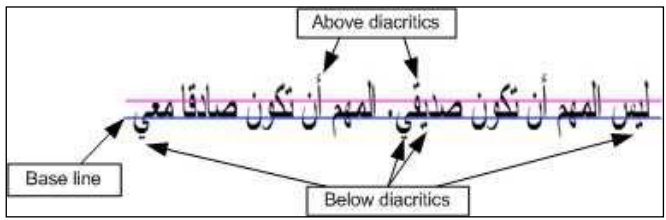

Figure 6. Sample of Arabic text showing some of its characteristics

Arabic script is written and read from right to left. The letters change shape presentation according to their position (at the beginning, middle or end of the word). Each letter can be written Arabic from 2 to 4 different forms *[12]*:

- When the letter is attached to any other: this is the position "isolated";
- When it is attached only to a letter from the front (left) is the position "initial";
- When it is attached only to a letter from behind (right): It is the position "final";
- And when it is attached at both ends at once: this is the position "median".

TABLE I. DIFFERENT FORMS OF ARABIC ALPHABETS

| Letter | isolated | Initial | Middle | Final |
|---|---|---|---|---|
| Alef | ا |  |  | ى ا ى |
| Ba' | ب | بـ | ـبـ | ـب |
| Ta' | ت | تـ | ـتـ | ت ة ـة |
| Tha' | ث | ثـ | ـثـ | ـث |
| Jeem | ج | جـ | ـجـ | ـج |
| H'a' | ح | حـ | ـحـ | ـح |
| Kha' | خ | خـ | ـخـ | ـخ |
| Dal | د |  |  | ـد |
| Thal | ذ |  |  | ـذ |
| Ra' | ر |  |  | ـر |
| Zai | ز |  |  | ـز |
| Seen | س | سـ | ـسـ | ـس |
| Sheen | ش | شـ | ـشـ | ـش |
| Sad | ص | صـ | ـصـ | ـص |
| Dhad | ض | ضـ | ـضـ | ـض |
| Tta | ط | ط | ـطـ | ـط |
| Dha' | ظ | ظ | ـظـ | ـظ |
| Ain | ع | عـ | ـعـ | ـع |
| Gahin | غ | غـ | ـغـ | ـغ |
| Fa' | ف | فـ | ـفـ | ـف |
| Qaf | ق | قـ | ـقـ | ـق |
| Kaf | ك | كـ | ـكـ | ـك |
| Lam | ل | لـ | ـلـ | ـل |
| Meem | م | مـ | ـمـ | ـم |
| Noon | ن | نـ | ـنـ | ـن |
| Ha' | ه | هـ | ـهـ | ـه |
| Waw | و |  |  | ـو |
| Ya' | ي | يـ | ـيـ | ـي |

### B. Existing Arabic OCR

Existing systems for the recognition of the Arabic script is based on the modeling of the three main entities: words, subwords and letters, respectively, and use all three types of approaches: global approach, pseudo-analytical approach and analytical approach *[9]*.

It is through the presentation of a few systems, as is the case for other records, that a comprehensive approach is still limited to a predefined vocabulary words and reduces target for a definite application (cheques, names of cities , and so on.). Systems based on such an approach have always entered into single words. These limits are also laid by Miled and al. *[18]*, and Ben Amara and al. *[1]* as part of their pseudo-analytical approaches based on the rules of concatenation of subwords known for recognizing words (names of cities).

We noticed that in the case of a large or open vocabulary and in the context of recognition of a text for instance, the systems are often based on an analytical approach. Within this framework, several problems were discussed, the most important being the segmentation of the text into words. This problem is more acute for writing Arabic for several reasons: there is an inter-word space and inter-subwords. Another major difficulty to report concerns the establishment of a post-processing because of a lack of vocabulary dictionaries and language tools easily integrated into systems of recognition.

We distinguish post-processing algorithms based on verification and correction by calculating distance editing in a dictionary (combinatorial methods) *[10]*, and the post Treatment based on the statistical structure of the language usually modelled by Hidden Markov Models (statistical methods) *[20] [Khorsheed,03]*.

In contrast to other languages such as English or French whose labels grammatical from a distributional approach characterized by a willingness to set aside any considerations relating to the meaning, the labels come from Arabic of an approach where alongside the formal semantics related to the morphology of the word, without reference to the latter's position in the sentence *[2] [3]*.

This is evidenced by the concepts of patterns and functions that occupy an important place in the grammar of Arabic. For example, the Arabic word "غَلَقَ" (close) is a verb in the 3rd person of the masculine singular accomplished active, however its form without vowels "غلق" admits grammatical four categories *[4]*:

- Substantive masculine singular "غَلْقٌ" (a closure)
- Word to the 3rd person of the masculine singular accomplished active "غَلَقَ"or "أغْلَقَ" (he closed)
- Word to the 3rd person of the masculine singular done "غُلِقَ" liabilities (it was closed)
- Word to the imperative 2nd person masculine singular "أغْلِقْ"(closed).

V. ARABIC VIDEO OCR

OCR is one of the most successful applications in the pattern recognition field. It is a common belief that OCR is a solved problem because so many papers and patents have claimed recognition rates as high as 99.99%. Although many commercial OCR systems work well on high quality scanned documents under controlled conditions, they fail in many tasks, such as video OCR, license plate OCR, and sign OCR. Current video OCR is limited to recognizing captions in video images for video indexing, or to identify license plates on vehicles for various applications *[5] [6]*.

The general model of Arabic video text recognition system can be described in terms of five stages: pre-processing, segmentation, feature extraction, classification, post-processing and recognition. Figure 7, illustrates these stages according to their order of occurrence.

The pre-processing stage is a collection of operations that apply some filters that enhance the text extracting from video frames thereby reducing noise and distortion, and consequently get better results from the recognition process. The segmentation stage decomposes the word into characters. The feature extraction stage analyzes a segment and selects a set of features that can be used to uniquely identify the text segment. The classification stage, which is the most the main stage in any OCR system, uses the features extracted in the previous stage to recognize the text segment. Finally, the post-processing stage, which improves the recognition by refining the decisions taken by the previous stage and recognizing words by using context.

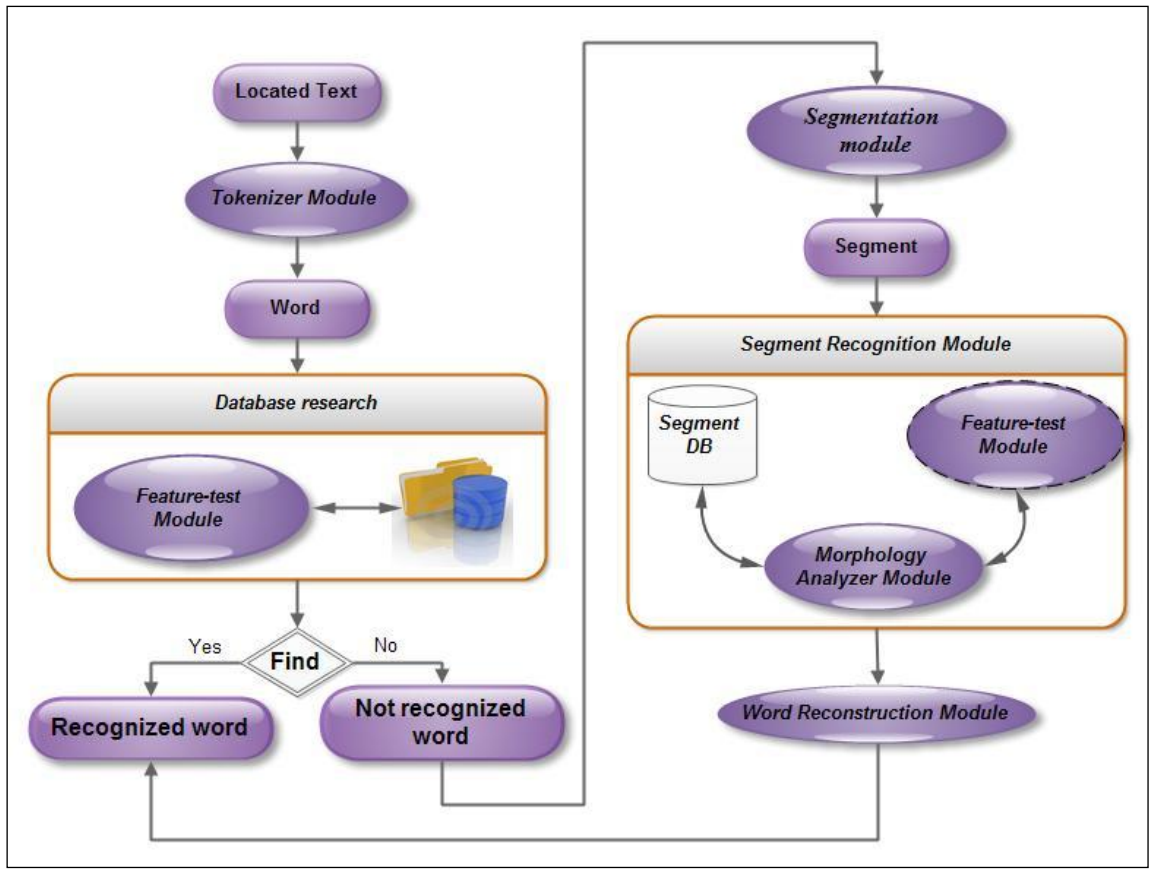

Figure 7. Different stages of proposed OCR system

A. *Pre-processing*

The recognition accuracy of OCR systems greatly depends on the quality of the input text and noise. The pre-processing stage attempts to compensate for poor quality originals and/or poor quality of Binarization stage after text extraction. This is achieved by reducing both noise and data variations.

1) *Text Enhancement*

We have chosen the method of Wolf, which uses the contents of all frames of an apparition of the same text to produce an enhanced image. This is done in a robust way, based on statistics calculated on the Gray level of each pixel during the time of onset. To automate this procedure we assume that the text appears only once.

We chose the bilinear interpolation, which calculates the gray level of a pixel as an average of gray levels of its neighbors. The weight of each neighbor is the distance to the pixel calculated:

$$p(x,y) = \frac{1}{\sqrt{(x-ix)^2 + (y-iy)^2}} \quad (2)$$

In different images, text may occur with various widths and heights. To have consistent features through all the text images and to reduce the variance of the size of the characters, the height of a text image is normalized. The height of the text image is scaled to a predefined size (26 pixels).

*2) Text Normalization*

In different images, text may occur with various widths and heights. To have consistent features through all the text images and to reduce the variance of the size of the characters, the height of a text image is normalized. The height of the text image is scaled to a predefined size (26 pixels).

The size of a character can vary from script to script, which can cause instability of the parameters (descriptors of the character). A natural technique of preprocessing is to bring the characters to the same size. We give in this section an algorithm to normalize the size. We begin by presenting a method of variation of the size of an image. Is an object of size $P \times Q$ we want to reduce the size $M \times N$.

The technique is as follows:

If $(i,j)$ is a pixel in the $M \times N$ final image, the corresponding pixel in the $P \times Q$ source image is:

$$\left(\left[\frac{(i*P)}{N}\right]\left[\frac{(j*Q)}{M}\right]\right) \quad (3)$$

The algorithm of size variation is:

> *1 - Let I and J the source image the final image.*
> *2 - for i = 0 to N-1*
> *     for j = 0 to M-1*
> *       J (i, j) = I ([(i * P) / N] [(j * Q) / M])*
> *     End for*
> *End for*

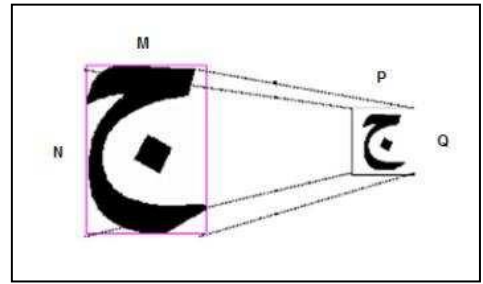

Figure 8. Example of character normalisation

*B. Tokenizer module*

All components that do not have pixels belonging to the baseline are considered diacritic. The baseline is the one that corresponds to the maximum of the horizontal projection. Information concerning the nature and location of these diacritical marks are registered for use in the future at the stage of recognition. The tokinizer module consists to isolates the words (tokens) of the located text. To build this part we have developed an algorithm that can identify and separate the punctuation marks as well as isolate any extra particles that may be attached to the beginning of the word while they are not part of it. A Database research module is used to check whether the database already contains the word. If the word is not present in the dictionaries, it will be segmented into character using a Segmentation module. Each segment will be recognized using a base of segments.

Detection of diacritics is done in two steps:

- The first step is to perform filtering of connected components based on fairly simple criteria: size of the bounding box, area, vertical layering. The aim is to reject most of diacritics, without rejecting the connected component corresponding to a body of a letter or pseudo-word.

- A second filter allows taking into account the fact that diacritics are either below or above the baseband.

After baseline detection (Figure 9) and diacritic elimination, a segmentation stage is necessary. This segmentation is based on a vertical projection (Figure 10). The segmenter should avoid cases of under-segmentation and/or over-segmentation. Each segment will be recognized using a base of segments.

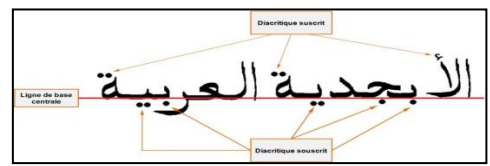

Figure 9. Example of base-line detection

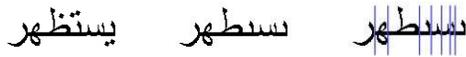

Figure 10. Example of text segmentation

The segmentation of the Arabic word into individual characters is a crucial step in recognizing Arabic text extracting from news sequences. Most of the recognition errors occur from segmentation errors. To minimize error, the segmenter, check before segmented, if there is an intersection pixel between the baseline and the letter to segment (T-junction). This condition gives the following decomposition of letters:

- The letter "ل" at the end of the word is considered as a whole segment.
- The letter "س" will be segmented into three elements.
- The letters "ب ق, ف, ث," will be segmented into two elements, by the segmenter, if they are at the end of the word.
- The letters "ص, ض" will be broken into two segments.

Other letters will be segmented as follows:

- The characters "ذ, د, ث, ب, ف, ث" will be detected by their sizes and positions relative to the subword. Indeed, they are always at the end of the word or subword. Such an error will be corrected by connecting the last segment, with its size, the previous segment.
- For the characters "ض, ص, ش, س", segments have generated a strong resemblance to the characters "ي, ن ت, ث,ب" at the beginning or middle of a word or subword. A segmentation fault will be corrected by testing diacritics above or below the segment. If the segment contains no diacritics, it is connected to the next segment. The problem here concerns the "-1" character. However, since the last segment of the character has no diacritic, it will be connected directly to the character that precedes it. And since an Arab character and is the most fragmented into three segments, the maximum number of connections should not exceed two connections.
- For the characters "ص, ش, س,ض ", the last segment obtained is similar to character" ن ". The difference lies in the above diacritic. A segmentation fault will be corrected by testing the diacritic mark. If this diacritic doesn't appear above the segment, (not the character "ن"), it will be connected to the previous segment.

*C. Feature extraction*

Once an OCR system has an isolated pattern (character or primitive), its next step is to extract the features of the pattern and then classify it. Feature extraction is one of the most difficult and important problems of pattern recognition. Different types and numbers of features may be extracted and used in the recognition stage. The selected set of features should be a small set whose values efficiently discriminate between patterns of different classes, but are similar for patterns within the same class. The feature extraction step is closely related to classification, because the type of features extracted here must match what the classifier expects.

A Feature-test module is used to get the token from the tokenizer module, get some information about it from the morphology analyzer module, and go through several tests one by one until it finds the part of speech of the word. A Morphology analyzer module is called by the feature-test module to analyze all words segments.

To illustrate the sequence of various stages of our system, we present in the following diagram the general synoptic system. The Features chosen to represent each character are:

- Extraction of occlusion: They correspond to the internal contour of the primary plots characters.
- The projection feature: In addition to the horizontal and vertical projection, we make a projection of the diagonal and the slanting diagonal. The projected result will be a vector of size equal to 160.
- Extraction of diacritics marks: diacritics are the secondary parts of characters. These marks above and below the characters, have an important role in the distinction of a few characters that differ only by the number or location of points. The following characters (اب ت ث -ﻟ) are differed only by the number and location of points.
- The transition features: the number of transition from 0 to 1 of the row, column, diagonal and slanting diagonal. The result is a vector of 160 features for each image of a letter.

*D. Classification*

Classification in an OCR system is the main decision making stage in which the extracted features of a test set are compared to those of the model set. Based on the features extracted from a pattern, classification attempts to identify the pattern as a member of a certain class. When classifying a pattern, classification often produces a set of hypothesised solutions instead of generating a unique solution.

Supervised classification methods can be used to identify a sample pattern. We used the fuzzy k-nearest neighbour algorithm (k=10). The coefficient of belonging of a new segment $x_i$ to class j is given by the formula:

$$u_{ij} = \frac{\sum_{t=1}^{k} u_{jt} \left( \|x_i - x_t\|^{m-1/2} \right)}{\sum_{t=1}^{k} \left( \frac{1}{\|x_i - x_t\|^{m-1/2}} \right)} \quad (4)$$

Where $u_{ij}$ is the coefficient of belonging to class $w_j$, the tth observation, among the k nearest neighbours of $x_i$. M is the variable determines the importance of the contribution of the distance in the calculating of the function of belonging.

*E. Post-processing*

The recognition rates of character recognition systems are not sufficient, hence is used to improve word recognition rate as opposed to character recognition rate. It is necessary to use context to detect errors and even to correct them. Post-processing is often implemented as a set of techniques that rely on character frequencies, lexicons, and other contextual information. As classification, sometimes, produces a set of possible solutions instead of a unique solution, post-processing is responsible for selecting the right solution using higher level information that is not available to the classifier. Post-processing also uses that higher level information to check the correctness of the solutions returned by the classifier. The post-processing operations are spell checking and correction. Spell checking can be as simple as looking up words in a lexicon.

In this stage a Word reconstruction module is used to gives words assumptions from segments previously recognized. The validation of alternative assumptions words is a user interactive job.

The post-processing stage, which is the final stage, improves recognition by refining the decisions taken by the previous stage and recognises words by using context. It is ultimately responsible for outputting the best solution, and is often implemented as a set of techniques that rely on character frequencies, lexicons, and other context information.

## VI. EXPERIMENTS

To validate our approach, we have used a varied database composed of news sequences extracted from different Arabic channels. Concerned channels are Tunisia 1 TV (tunisiatv.com) as generalist channel presenting news at 13H and 20H. We have also tried our approach with news video extracted from Aljazeera (Aljazeera.net) and AlArabiya (alarabiya.net) which are specialist channels presenting news continually.

*A. Experimental results of extraction text*

To experiment our previous work we have based on the number of identified correct text contours, we find the following recall and precision rate to evaluate the extraction text method. Figure 11 shows the results of this experiment.

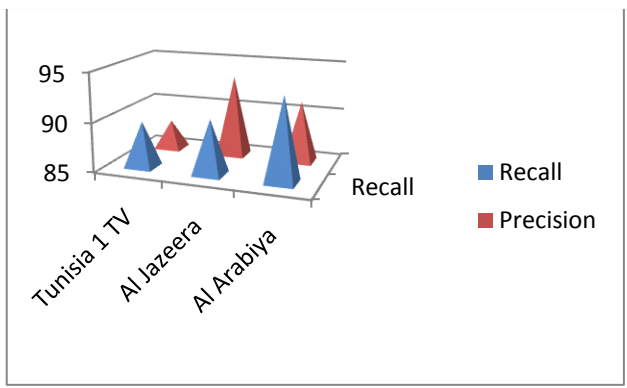

Figure 11. Evaluation results of textual localization

We notice that Al Jazzeera and Al Arabiya channels present the best rate of recall and precision because they present the best quality of graphic text. In fact, our approach is more robust to various font size, font styles, contrast levels and background complexities because it uses both of colour features and edges features to differentiate text pixels from background pixels, besides it's based on a neural network trained on different types of text styles.

*B. Experimental results of Recognition text*

In order to investigate the effectiveness of our recognition sub-system in recognizing Arabic Text extracted from various news sequences, a series of tests were performed using fuzzy KNN classifier with K=3, K=5 and K=10. Figure 12 shows the recognition results.

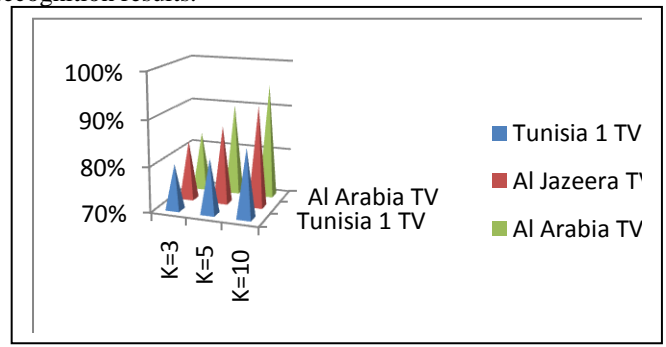

Figure 12. Evaluation results of text recognition

We note that the rate of recognition of texts extracted from Al-Arabiya and Al-Jazeera TVs are better than those extracted from Tunisia 1 TV. This is explained by the fact that the text extract from Al-Arabiya and Al-Jazeera TVs is clearer and more readable. We find it is difficult to recognize small size characters that appear frequently in news videos. The best recognition rate is in all cases for K=10.

This recognition system was also tested on the database "Houda"[17]as part of "ICDAR 2009 Handwritten Farsi/Arabic Character Recognition Competition" [19]. Hoda database which contains handwritten digits is presented by Khosravi and Kabir in 2007 [Khosravi, 2007]. Binary images of 102,352 digits were extracted from about 12,000 registration forms. These forms were scanned at 200 dpi with a high speed scanner. Hoda database was partitioned into train (60,000 samples) and test (20,000 samples) subsets.

## VII. CONCLUSION

In this paper, we propose an efficient method to deal with background complexity in Arabic news video text recognition by efficiently integrating multiple frame information. Proposed system includes several tasks such as Text Detection; Localization; Segmentation; and Recognition. By using this system we can produce quite clear text for Arabic text recognition. The extraction rate has been increased about 91% and 84% for text recognition. These methods can also be adopted by any other type of Video OCR systems to increase recognition rate.

Used OCR techniques do not perform very well when the extracted characters have a too low resolution. The development of new OCR techniques to recognize low resolution characters is still necessary. Other aspect is computation reduction for mobile image text recognition applications. Most mobile devices, such as mobile phones, have less computation power and less memory resources than a desktop computer. In order to build a video text extraction application on these devices, the algorithms proposed in this paper need to be optimized or even modified to reduce the computation cost.

ACKNOWLEDGMENT

The authors would like to acknowledge the financial support of this work by grants from General Direction of Scientific Research (DGRST), Tunisia, under the ARUB program.